# Estimation of Inter-Sentiment Correlations Employing Deep Neural Network Models

Xinzhi Wang
Tsinghua University, Beijing, China
wxz15@mails.tsinghua.edu.cn
Carnegie Mellon University, Pittsburgh,
U.S. xinzhiw@andrew.cmu.edu

Shengcheng Yuan
Rutgers University, Newark, New
Jersey, U.S.
shengcheng.yuan@rutgers.edu

Hui Zhang *
Tsinghua University, Beijing, China
zhhui@mail.tsinghua.edu.cn

Yi Liu
People's Public Security University,
China
liuyi789@163.com

## ABSTRACT

This paper focuses on sentiment mining and sentiment correlation analysis of web events. Although neural network models have contributed a lot to mining text information, little attention is paid to analysis of the inter-sentiment correlations. This paper fills the gap between sentiment calculation and inter-sentiment correlations. In this paper, the social emotion is divided into six categories: *love, joy, anger, sadness, fear, and surprise*. Two deep neural network models are presented for sentiment calculation. Three datasets – the titles, the bodies, the comments of news articles – are collected, covering both objective and subjective texts in varying lengths (long and short). From each dataset, three kinds of features are extracted: explicit expression, implicit expression, and alphabet characters. The performance of the two models are analyzed, with respect to each of the three kinds of the features. There is controversial phenomenon on the interpretation of *anger* (fn) and *love* (gd). In subjective text, other emotions are easily to be considered as *anger*. By contrast, in objective news bodies and titles, it is easy to regard text as caused *love* (gd). It means, journalist may want to arouse emotion *love* by writing news, but cause *anger* after the news is published. This result reflects the sentiment complexity and unpredictability.

## 1. INTRODUCTION

Social sentiments are valuable. Effective sentiment calculation does not only help product developers to understand the preference of their customers, but also helps governors to evaluate the public opinion to a social event. In recent years, machine learning algorithms have contributed a lot to sentiment analysis in text. However, even though the performance of these algorithms has been improved a lot, little attention is paid to the complexity of social sentiments: in social psychologist's views, social sentiments are compound and diverse. For instance, someone could say, "You stupid farmer, why did you save the snake when you knew he could kill you?", where a strong sentiment of empathy can be easily confused with anger. Given the same social event through text, the sentiments of readers may diverse. People with different personalities and diversified prior knowledge tend to pay attention to different aspects. If a reader takes care of more than one aspect of the sentiment, then his/her words may carry more than one kind of sentiment. In the field of psychology, there is still some controversy about the classification of sentiments. However, in engineering scientist's views, the learning algorithms often treat the social sentiments as a simple classification (e.g., positive/negative). The social psychologists and the engineering scientists focus on their own perspectives of sentiment calculation, leaving a gap between the performance of sentiment classification and the analysis of sentiment correlations. To fill this gap, this paper focuses on six kinds of human emotions, i.e., love, joy, anger, sadness, fear, surprise[34], builds two deep learning models, and extracts three different kinds of features, to support mining both the text sentiment and the inter-sentiment correlation. The above six categories of sentiments are widely accepted today [34]. If other sentiment categories are chosen, the analysing process can be analogized.

Essentially, the performance of a sentiment calculation model is affected by both the quality of data and the effectiveness of model. Just as described above, the public's sentiments to a social event are compound. On one hand, the sentiment cognition is heterogeneous among people even to the same text. It means there isn't one hundred percent accurate data for each person. Consequently, a better understanding of sentiment data can help interpreting the performance of sentiment calculation models. On the other hand, the sentiment calculation result could provide clues for data understanding, improving the comprehension of inter-sentiment correlations. To understand the sentiment data, three datasets are collected, covering both objective and subjective texts in varying lengths (long and short). From each dataset, three kinds of features are extracted: explicit expression, implicit expression, and alphabet characters. To calculate sentiment for both the short and long texts uniformly, this paper presents two deep neural network models.

The rest of the paper is organized as follows. We start by reviewing related work. In Section 3, we describe the sentiment calculation models and the inter-sentiment correlation analysis method. In Sections 4, the details of the experiment are introduced. Finally, we draw conclusions in Section 5.

## 2. RELATED WORK

Social media is playing increasingly important roles in scientific research as well as our daily life. The data from social media contribute to the improvement of text-related analyses, such as sentiment analysis [3, 7, 23], sarcasm detection [1, 2], event dissemination [3], short text clustering [4], user clustering [5, 6], knowledge recommendation [7], and user behaviour analysis [8, 9]. Continuous word representations, including word2vec [10], glove [11], and weighted word embedding [12], also provide new ideas on knowledge mining.

Sentiment analysis, as an important branch of knowledge mining, can be categorized into three levels, namely word level, sentence level, and article level. In word level, sentiment words are extracted mainly through three ways: 1) manual approach [13], 2) dictionary-based approach [14, 15], and 3) corpus-based approach [16]. In sentence level, intra-sentential and inter-sentential sentiment consistency were explored [17]. Qiu et al. [18] employed dependency grammar to describe relations for double propagation between features and opinions. Ganapathibhotla and Liu [19] adopted dependency grammar for sentiment analysis of comparative sentences. The Conditional Random Fields (CRF) method [20] was used as the sequence learning technique for extraction. Machine learning methods are widely used in both sentence and article level. Naive Bayesian [21, 22], maximum entropy classification [22], Support Vector Machines (SVMs) [22], and pattern recognition methods [23] are employed frequently. In recent years, neural network models, such as Long Short Term Memory (LSTM) [24, 25], convolutional neural network (CNN) [26, 27], recursive auto-encoders [28, 29], adversarial learning [30] and attention mechanism [31, 32], also contribute to sentiment analysis and classification tasks.

All the above works have improved the performance of sentiment analysis. However, just as Wilson et al. [33] pointed out, a single text may contain multiple opinions. Parrott [34] demonstrated that human sentiments are prototyped and complex. Most of the recent works just focus on recognizing the sentiment expressed in text. Little attention is paid to associate sentiment calculation in engineering with inter-sentiment correlation in psychology.

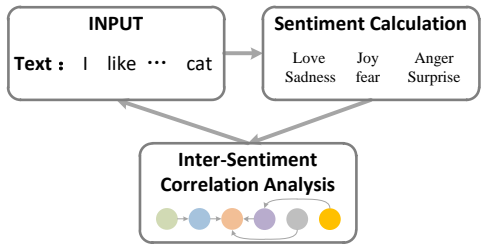

**Figure 1: Interaction diagram of sentiment calculation and inter-sentiment correlation analysis**

## 3. MODEL

Sentiment data and model are the two factors which influence the performance of sentiment analysis. The compound sentiment in data influences the result of the sentiment analysis model, and the model's result furthermore impacts analysis of the inter-sentiment correlation. The interaction process is shown in Fig. 1. This section introduces the sentiment calculation models and the inter-sentiment correlation analyzation method separately. Namely, sentiment analysation is divided into two parts in this paper. The first is sentiment calculation, the second is inter-sentiment correlation analyzation.

### 3.1 Sentiment Calculation Models

The sentiment calculation models aim to discriminate the sentiment orientation of input texts. In this paper, two deep neural network models, CNN-LSTM2 and CNN-LSTM2-STACK, are employed to calculate sentiment. In both models, the length of an input text can be either short or long. The output of the models is one of the six kinds of emotions, i.e., *love*, *joy*, *anger*, *sadness*, *fear*, and *surprise*. The calculation process can be divided into three parts, as shown in Fig. 2. CNN-LSTM2 is constructed with Part I and Part II. CNN-LSTM2-STACK is constructed by adding an additional Part III to CNN-LSTM2. The details of the three parts are represented as follows.

*3.1.1 Part I: Feature Processing.* Part I focuses on feature processing which transforms the original features into dense vector information. There are four operations in this part: vector lookup, window sliding, convolutional calculation, and ReLU activation.

Let the input is denoted as $w_i^j$, which means the *i*-th feature of the *j*-th sample text. The *j*-th sample text is indicated as $[w_1^j, w_2^j, ..., w_N^j]$, where the text is padded by 'none' to length of $N$. Here, 'none' is the reserved symbol in the vocabulary. For instance, if the first sample text is 'I like small cat' and $N = 5$, then $w_1^1 = 'I'$, $w_2^1 = 'like'$, $w_3^1 = 'small'$, $w_4^1 = 'cat'$, $w_5^1 = 'none'$.

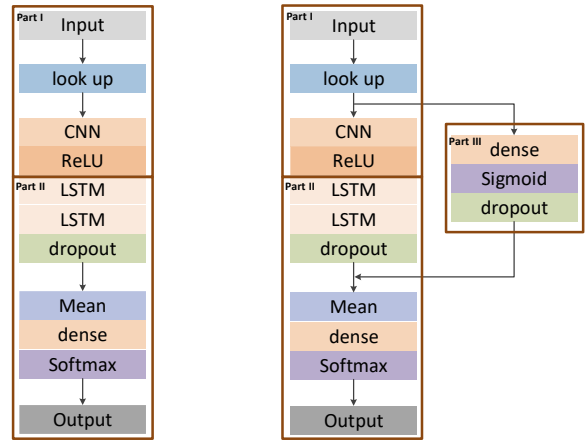

(a) CNN-LSTM2    (b) CNN-LSTM2-STACK

**Figure 2: Sentiment calculation models**

The first operation, vector lookup, searches the embedding representation $v_i^j$ of the corresponding input feature $w_i^j$, which is formulated as:

$$v_i^j = \text{LOOKUP}(w_i^j) \qquad (1)$$

The second operation, window sliding, packets a targeted input feature and its context together after 'none' padding. Specifically, the window size in this paper is set to be 5, and $w_0^j = w_{-1}^j = $ 'none'. Then, in the third operation, a convolutional layer is applied.

$$h_{i,1}^j = \text{CNN}([v_{i-2}^j, v_{i-1}^j, v_i^j, v_{i+1}^j, v_{i+2}^j]) \qquad (2)$$

where $h_{i,1}^j$ is the first hidden layer. And $[.]$ means embedding concatenation. In the last operation, a ReLU activation layer is added.

$$h_{i,2}^j = \text{ReLU}(h_{i,1}^j) \approx \log(1 + h_{i,1}^j) \qquad (3)$$

where $h_{i,2}^j$ is the second hidden layer of the $i$-th feature of the $j$-th sample. $h_{i,2}^j$ acts as the input of Part II.

*3.1.2 Part II: Sentiment Calculation.* Part II focuses on the sentiment calculation after feature processing of Part I. There are five operations in this part, i.e., Long Short Term Memory (LSTM) calculation, dropout operation, average calculation, fully connect calculation, and sorftmax.

Firstly, the $h_{i,2}^j$ output of Part I is fed into a two-layer LSTM component. The outputs of the two LSTM layers are represented as $h_{i,3}^j$ and $h_{i,4}^j$ respectively. After that, the dropout operation is applied to prevent over-fitting. Then, we can get:

$$h_{i,3}^j = \text{LSTM}(h_{i,2}^j) \qquad (4)$$

$$h_{i,4}^j = \text{LSTM}(h_{i,3}^j) \qquad (5)$$

$$h_{i,5}^j = \text{DROPOUT}(h_{i,4}^j) \qquad (6)$$

where $i$ is the index of the text sequence, which is an integer ranging from 1 to N.

In practice, even though the texts have been padded to the same length of $N$, their actual lengths still vary. To settle this problem, we define $ms_i^j \in \{0,1\}$ as mask. The sequence data is combined to fixed-length vector.

$$h_{\_,6}^j = \frac{1}{N}\sum_i(h_{i,5}^j \cdot ms_i^j) \qquad (7)$$

If the $i$-th feature of the $j$-th sample text is valid (i.e., not "none"), then $ms_i^j = 1$. Otherwise, $ms_i^j = 0$. Note that $h_{\_,6}^j$ denotes the sixth hidden layer. The last two steps are a fully connected layer and a softmax layer.

$$h_{\_,7}^j = \text{LINEAR}(h_{\_,6}^j) = W^T h_{\_,6}^j + b \qquad (8)$$

$$P(e_l^j) = \text{SOFTMAX}(h_{\_,7}^j) = \frac{e^{h_{\_,7,l}^j}}{\sum_{m=1}^M e^{h_{\_,7,m}^j}} \qquad (9)$$

where W and $b$ are the weight matrix and bias. $l$ is the sentiment index which ranges from 1 to 6, corresponding to the six categories of sentiments.

*3.1.3 Part III: Original Feature Attention.* Part I and Part II construct the model CNN-LSTM2. However, with the neural network going deep, the backward fine-tuning process in CNN-LSTM2 becomes weak, and the vanishing gradient problem occurs. To solve this problem, a second model CNN-LSTM2-STACK is constructed by associating CNN-LSTM2 with Part III. This part links the input feature embedding $v_i^j$ to the layer $h_{\_,6}^j$ through linear and sigmoid operations.

$$h_{i,8}^j = \text{LINEAR}(v_i^j) \qquad (10)$$

$$h_{i,9}^j = \text{SIGMOID}(h_{i,8}^j) \qquad (11)$$

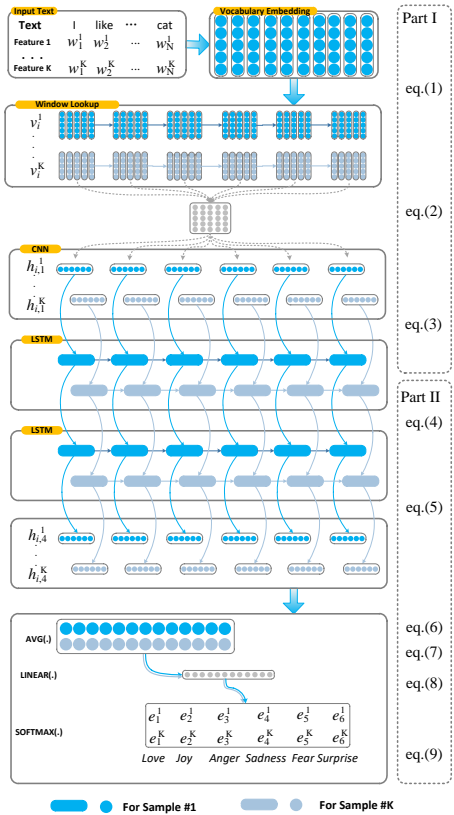

**Figure 3: Model of CNN-LSTM2**

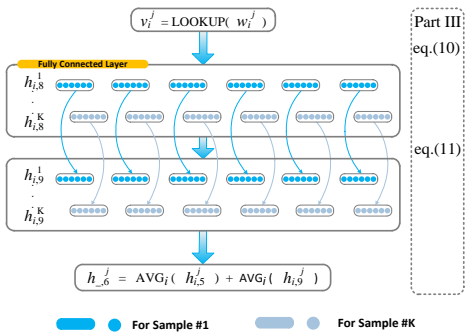

**Figure 4: Model of CNN-LSTM2-STACK**

Then, layer $h^j_{\_,6}$ is adjusted by $h^j_{i,9}$, i.e., the inputs of the sixth hidden layer changes to both $h^j_{i,5}$ and $h^j_{i,9}$.

$$h^j_{\_,6} = \frac{1}{N}\Sigma_i(h^j_{i,5} \cdot ms^j_i) + \frac{1}{N}\Sigma_i(h^j_{i,9} \cdot ms^j_i) \quad (12)$$

The other operations in Part II remain the same. Part III aims to emphasize the impact of the input feature embedding on the sentiment calculation result. In other words, by stacking Part III, the network pays more attention to the original feature information.

The two models CNN-LSTM2 and CNN-LSTM2-STACK use deep neural network methods to calculate text sentiment. In our model design, both long and short texts can act as input. Three parts – feature processing, sentiment calculation, and original feature attention – are introduced. The details are shown in Fig. 3 and Fig. 4. The sentiment calculation result of the two models supports the inter-sentiment analysis below.

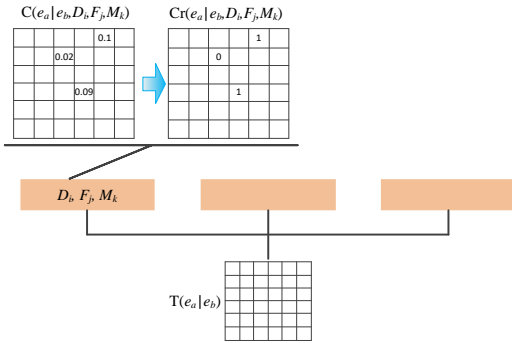

**Figure 5: Process of sentiment correlation analysis**

## 3.2 Inter-Sentiment Correlation Analysis

In terms of the content, sentiment data can be divided into two categories: objective texts and subjective texts. Objective texts tell the story of what happens in an event. Subjective texts are the comments from readers, which are aroused by the objective texts. In this paper, the titles and the bodies of news articles are collected as objective texts, which are edited and published by journalist. The comments of news articles are collected as subjective texts, which are generated by the public. Analysis of objective texts aims at finding the correlation of objective information with subjective sentiment. Analysis of subjective texts intends to mine the correlation of subjective information to sentiment cognition.

Suppose E is the input sentiment label of a sample text, which is marked manually. $\widetilde{E}$ is the ground truth of the sentiment, and $\widehat{E}$ is the output label predicted by a learning model. If $\widetilde{E} = \widehat{E} = E$, it means our model's prediction, the manually marked label, and the ground truth of the sentiment are in consistence. However, it is almost impossible to completely keep this consistence in all the sample texts in practice. On one hand, the learning model often has a generalization error rate, which generates the differences between $\widehat{E}$ and E. On the other hand, the difference between E and $\widetilde{E}$ stems from the human sentiment cognitive bias. The sentiment cognitive bias restricts the quality of data and impacts on the prediction in further steps. To estimate the above two factors that influence the consistence, a voting mechanism based on multiple models and datasets is put forward, as shown in Fig. 5.

Let $C(e_a|e_b, D_i, F_j, M_k) \in [0,1]$ represent the probability that $e_b$ is recognized as $e_a$, given data $D_i$, feature $F_j$ and model $M_k$. Let $Cr(e_a|e_b, D_i, F_j, M_k) \in \{0,1\}$ denote the indicator of $e_b$ being recognized as $e_a$, where $Cr$ is calculated by the following equations:

$$\begin{cases} Cr(e_a|e_b, D_i, F_j, M_k) = \{C(e_a|e_b, D_i, F_j, M_k) > \theta\} \cdot \{1\} \\ \{True\} \cdot \{1\} = 1 \\ \{False\} \cdot \{1\} = 0 \\ \theta \in [0,1] \end{cases} \quad (13)$$

where $D_i, F_j, M_k$ indicate corresponding dataset, input feature, and model respectively. $\theta$ is a threshold. $Cr$ is the sentiment correlation matrix with elements belonging to $\{0, 1\}$. The voting process can be formulated as:

$$T(e_a|e_b) = \wedge_{i,j,k} Cr(e_a|e_b, D_i, F_j, M_k) \quad (14)$$

The voting result of objective texts (i.e., $T(e_a|e_b)$ for news titles and bodies) implies the correlations from objective information to subjective sentiment. The voting result of subjective texts (i.e., $T(e_a|e_b)$ for news comments) give clues on the correlations from subjective information to sentiment cognition. The result is shown in the experiment section.

## 4. EXPERIMENT

### 4.1 Datasets

The datasets used in this paper is crawled from one of the most popular social network, news channel (http://news.sina.com.cn/society/moodrank/). Each news article is split into three parts: the comment (Data #1), the body (Data #2), and the title (Data #3), where Data #1 is treated as subjective text, and Data #2 and Data #3 are regarded as objective text. The sentiment labels of the three datasets are generated through the vote of the public, strong rules, and manual selection. The distribution of the sentiments in the three datasets is introduced in Table 1. The comment dataset contains more than 150,000 sample texts. News body and news title data both contain more than 24,000 sample texts.

### 4.2 Data Feature and Model Parameters

A text is composed of words, words are composed of characters, and character is one of the most basic features of text. This paper presents three different ways to extract features from a sample text. They are explicit expression, implicit expression, and character features. In explicit expression, the features are all the words in the text. In implicit expression, the features are the synonym tags of the words, where the synonym tags are extracted through HIT synonymous dictionary (HIT IR-Lab Tongyici Cilin (Extended)). It means synonyms in the dictionary share the same symbol. In character features, each letter (or character) in the text is an independent feature. Table 2 shows the statistical information of features in the three datasets. The number of features, the maximum, minimum and average number of features per document are listed in the table.

**Table 1: Data Distribution of Sentiments**

| Sentiment | Data #1 News comment | | Data #2 News body | | Data #3 News title | |
|---|---|---|---|---|---|---|
| | Train | Test | Train | Test | Train | Test |
| *Love* | 29,658 | 7,398 | 6,788 | 1,698 | 6,717 | 1,680 |
| *Fear* | 2,098 | 520 | 5,240 | 1,309 | 5,211 | 1,302 |
| *Joy* | 15,426 | 3,848 | 5,178 | 1,295 | 5,159 | 1,290 |
| *Sadness* | 9,238 | 2,304 | 1,457 | 364 | 1,451 | 364 |
| *Surprise* | 13,283 | 3,311 | 323 | 82 | 272 | 68 |
| *Anger* | 51,610 | 12,882 | 578 | 144 | 571 | 142 |
| **Total** | **121,313** | **30,263** | **19,564** | **4,892** | **19,391** | **4,846** |

**Table 2: Statistical Information of Features in the Datasets (Fea: feature; Exp: explicit expression; Imp: implicit expression; Char: Character features; #Fea: the number of features; #f/doc: maximum, minimum, and average number of features per document)**

| Fea | Data #1 News comment | | Data #2 News body | | Data #3 News title | |
|---|---|---|---|---|---|---|
| | #Fea | #f/doc | #Fea | #f/doc | #Fea | #f/doc |
| Exp | 81,519 | 1,553/1/21 | 155,327 | 8,297/4/718 | 20,643 | 1,936/6/13 |
| Imp | 55,058 | 1,553/1/21 | 86,057 | 8,297/4/718 | 14,585 | 1,936/6/13 |
| Char | 5,707 | 2,529/2/31 | 5,939 | 12,493/8/1,164 | 3,327 | 1,937/12/21 |

In both the sentiment calculation models, the dimension of embedding is set to be 100, of which the initial value is assigned randomly. The output dimension of the convolutional layer is set to 100 and the output dimension of LSTM layers are set to 128. The dimension of the stack layer (i.e., Part III) is also set to 128. The final output dimension of the model is 6, corresponding to the six categories of emotions.

## 4.3 Sentiment Calculation Result

Considering the two models (CNN-LSTM2 and CNN-LSTM2-STACK), the three kinds of features (explicit expression, implicit expression and character), and the three datasets (comments, bodies and titles) together, there are 18 (2 × 3 × 3) combinations of choices in total to set up the sentiment calculation. The performance of the sentiment calculation on all the combinations are observed. During the training process, cross entropy is employed as the loss function. The loss and accuracy trends are recorded. Precision, recall, and F1-score on test sets are also recorded for all the sentiment labels in each training epoch. The tags listed in Table 3 are used to presents the six sentiment categories. We use 'tag'-precision, 'tag'-recall, and 'tag'-f1 to represent sentiment precision, sentiment recall, and sentiment F1-score respectively. Besides, the accuracy on test data is also recorded as shown in Fig. 6, Fig. 8 and Fig. 10.

In this section, we firstly show the results grouped by dataset. A comprehensive discussion combining all the results is given in Section 4.4.

**Table 3: Tags of Sentiments**

| Sentiment | Tag | Sentiment | Tag |
|---|---|---|---|
| *Love* | gd | *Sadness* | ng |
| *Fear* | zj | *Surprise* | xq |
| *Joy* | gx | *Anger* | fn |

*4.3.1 Data #1: News Comments.* In Data #1, among all the combinations, model CNN_LSTM2 with the explicit expression features rank the top on accuracy (85.0%). If given the same model (either CNN_LSTM2 or CNN_LSTM2_STACK), the explicit expression features always perform the best in all the three kinds of features, followed by the character features. It is also shown that given the same features, model CNN_LSTM2 performs better on accuracy than model CNN_LSTM2_STACK does. The details are shown in Fig. 6 and Table 4.

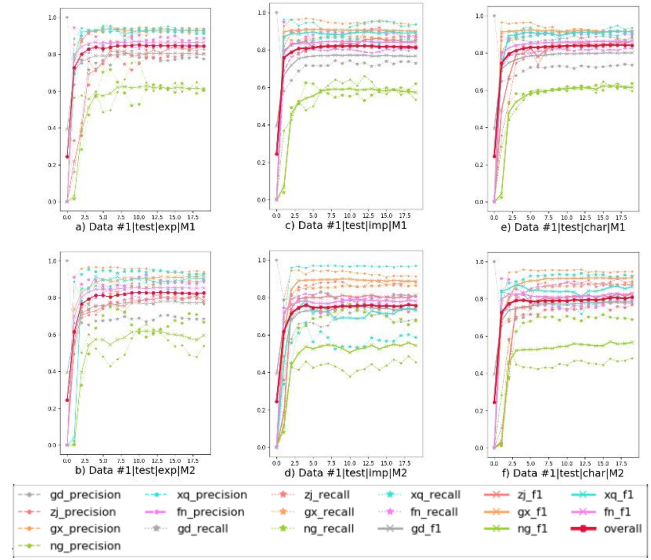

**Figure 6: Precision, recall, and F1-score trend on test set of Data #1**

The precision, recall, and F1-score of sentiment *joy* (gx) performs well. *Sadness* (ng) performs badly on both models. This phenomenon also reflects that sentiment *sadness* is more difficult to be recognized than other sentiments are, for example, the sentiment in the text 'so happy to cry'.

The accuracy and loss trends on training set of Data #1 are shown in Fig. 7. In every combination of the features and the models, the accuracy becomes steady after several iterations and achieves larger than 90%. The accuracy of both explicit expression features and implicit expression features reach to higher than 97% on the two models. The accuracy of character features on CNN_LSTM2 and CNN_LSTM2_STACK reach to 91.63% and 92.52% respectively, and the corresponding loss of the two models are 22.96% and 21.69%.

All the three kinds of features (i.e., explicit expression, implicit expression, and character) are effectively fitted by both models. The fitting result of explicit expression and implicit expression is slightly better than that of the character features on the training set. However, the fitting result on the test set shows an overfitting problem for explicit expression and implicit expression on the two models. Character features perform relatively consistent in both the training and testing sets.

**Table 4: Accuracy Rank on Test Set of Data #1**

| Rank | Feature | Model | Accuracy |
|---|---|---|---|
| 1 | Explicit | CNN_LSTM2 | 85.0% |
| 2 | Character | CNN_LSTM2 | 84.4% |
| 3 | Explicit | CNN_LSTM2_STACK | 82.9% |
| 4 | Implicit | CNN_LSTM2 | 82.2% |
| 5 | Character | CNN_LSTM2_STACK | 81.5% |
| 6 | Implicit | CNN_LSTM2_STACK | 76.1% |

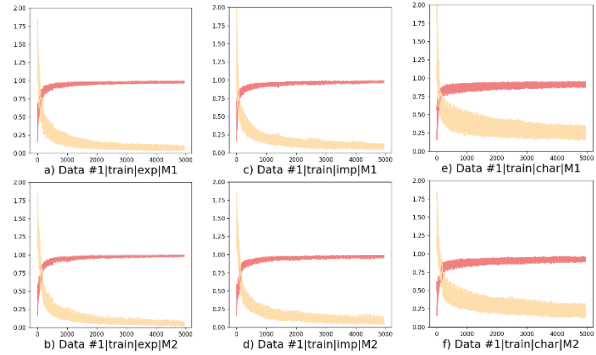

**Figure 7: Accuracy and loss trends on training set of Data #1**

*4.3.2 Data #2: News Bodies.* In Data #2 (news bodies which is kind of long text), model CNN_LSTM2_STACK with explicit expression features achieves the best score on accuracy (82.5%) among all the six combinations of models and features. In terms of the features, it is shown that explicit expression features always perform the best, no matter which model is applied. Implicit expression features give a little lower accuracy than explicit expression features. However, character features fail to achieve a good score on accuracy in both models. The result is shown in Fig. 8 and Table 5.

Precision, recall, and F1-score of sentiment *love* (gd) perform well, while those of *anger* (fn), *surprise* (xq), and *sadness* (ng) perform bad. It shows differences in performance among sentiments. We will discuss the reasons for this phenomenon in Section 4.4. In terms of the features, explicit expression features perform best, while character features perform worst, which indicates that character features are not suitable for long texts in our models.

**Table 5: Accuracy Rank on Test Set of Data #2**

| Rank | Feature | Model | Accuracy |
|---|---|---|---|
| 1 | Explicit | CNN_LSTM2_STACK | 82.5% |
| 2 | Explicit | CNN_LSTM2 | 81.2% |
| 3 | Implicit | CNN_LSTM2 | 79.6% |
| 4 | Implicit | CNN_LSTM2_STACK | 79.1% |
| 5 | Character | CNN_LSTM2 | 62.0% |
| 6 | Character | CNN_LSTM2_STACK | 55.3% |

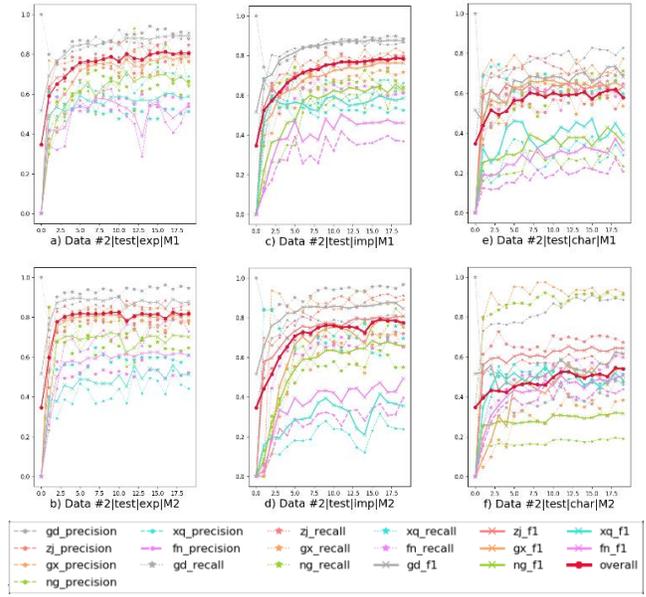

**Figure 8: Precision, recall, and F1-score trend on test set of Data #2**

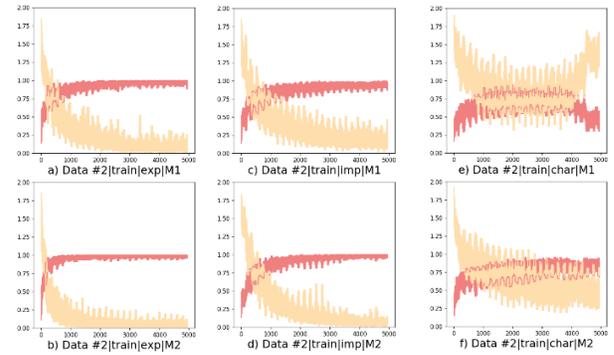

**Figure 9: Precision and loss trends on training set of Data #2**

The accuracy and loss trends on training set of Data #2 are shown in Fig. 9. The accuracy and loss of both implicit expression and explicit expression features becomes steady after several iterations. The accuracy of these two kinds of features on model CNN_LSTM2_STACK achieves higher than 98%, and the loss reaches to 0.03 and 0.05 respectively. On model CNN_LSTM2, the accuracy of these two features are 79% and 85% respectively, and loss are 0.68 and 0.49 respectively. From the above results, it can be concluded that model CNN_LSTM2_STACK fits better than model CNN_LSTM2. The accuracy of character features on the two models cannot converge, which means character features are not suitable for long text process, which is in line with the former description.

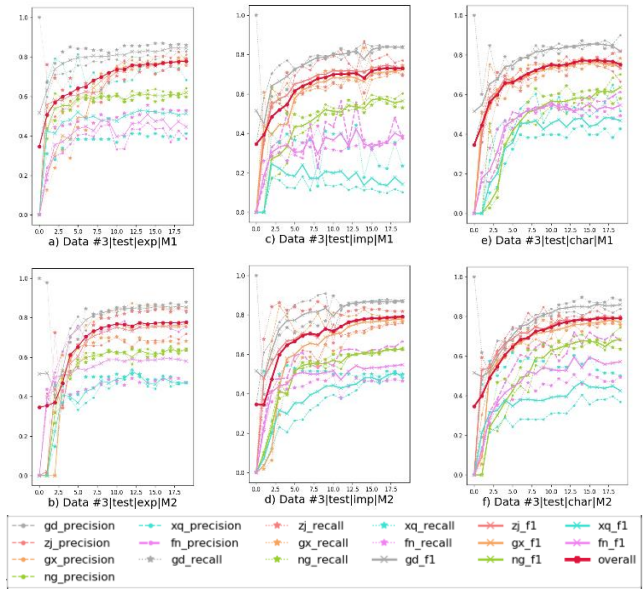

Figure 10: Precision, recall, and F1-score trend on test set of Data #3

*4.3.3 Data #3: News Titles.* In Data #3, all the six combinations of the models and the features give relatively similar scores of accuracies, ranging from 77.8% to 82.0%. Model CNN_LSTM2_STACK combined with the character features performs the best, with an accuracy of 82.0%. Model CNN_LSTM2_STACK performs better than model CNN_LSTM2, no matter which kind of features is chosen. In terms of the precision, recall and F1-scores in the six sentiments, *love* (gd) performs the best, while *anger* (fn) and *surprise* (xq) perform the worst. More details are shown in Fig. 10 and Table 6.

The result of Data #3 is consistent with the result of Data #2. The objective texts with the ground truth of *anger* (fn) or *surprise* (xq) can confuse the models, making the models incorrectly classify the texts which belong to *anger* (fn) or *surprise* (xq) into other sentiment categories.

The accuracy and loss trend on training data are shown in Fig. 11. The accuracy of all the three kinds of features on the two models becomes steady after iterations. Among them, model CNN_LSTM2 combined with implicit expression performs the worst, achieves an accuracy of 94% and a loss of 0.16. The accuracy of the other five groups are higher than 97% and the losses are less than 0.1.

**Table 6. Accuracy Rank on Test Set of Data #3**

| Rank | Feature | Model | Accuracy |
|---|---|---|---|
| 1 | Character | CNN_LSTM2_STACK | 82.0% |
| 2 | Implicit | CNN_LSTM2_STACK | 81.2% |
| 3 | Explicit | CNN_LSTM2_STACK | 80.5% |
| 4 | Explicit | CNN_LSTM2 | 80.0% |
| 5 | Character | CNN_LSTM2 | 79.6% |
| 6 | Implicit | CNN_LSTM2 | 77.8% |

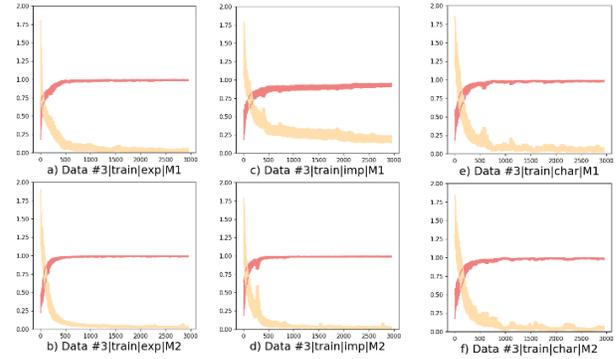

Figure 11: Precision and loss trends on training set of Data #3

Table 7: The Stable Sentiment Calculation Results (on test sets) of Three Features of Three Datasets on Two Models. (D: data; M1: model CNN_LSTM2; M2: CNN_LSTM2_STACK; A: Accuracy)

| D #1 | gd_f1 | zj_f1 | gx_f1 | ng_f1 | xq_f1 | fn_f1 | A |
|---|---|---|---|---|---|---|---|
| exp\|M1 | 0.804 | 0.796 | 0.926 | 0.622 | 0.928 | 0.869 | **0.850** |
| exp\|M2 | 0.778 | 0.803 | 0.913 | 0.619 | 0.886 | 0.855 | 0.829 |
| imp\|M1 | 0.772 | 0.855 | 0.903 | 0.588 | 0.895 | 0.844 | 0.822 |
| imp\|M2 | 0.738 | 0.796 | 0.886 | 0.558 | 0.741 | 0.785 | 0.761 |
| char\|M1 | 0.803 | 0.842 | 0.917 | 0.620 | 0.917 | 0.864 | 0.844 |
| char\|M2 | 0.794 | 0.820 | 0.915 | 0.586 | 0.842 | 0.837 | 0.815 |
| **D #2** | **gd_f1** | **zj_f1** | **gx_f1** | **ng_f1** | **xq_f1** | **fn_f1** | **A** |
| exp\|M1 | 0.888 | 0.808 | 0.789 | 0.699 | 0.601 | 0.517 | 0.812 |
| exp\|M2 | 0.890 | 0.816 | 0.812 | 0.709 | 0.558 | 0.616 | **0.825** |
| imp\|M1 | 0.876 | 0.800 | 0.772 | 0.660 | 0.553 | 0.531 | 0.796 |
| imp\|M2 | 0.868 | 0.803 | 0.787 | 0.666 | 0.417 | 0.473 | 0.791 |
| char\|M1 | 0.736 | 0.606 | 0.654 | 0.398 | 0.450 | 0.372 | 0.620 |
| char\|M2 | 0.668 | 0.634 | 0.566 | 0.312 | 0.494 | 0.497 | 0.553 |
| **D #3** | **gd_f1** | **zj_f1** | **gx_f1** | **ng_f1** | **xq_f1** | **fn_f1** | **A** |
| exp\|M1 | 0.851 | 0.817 | 0.802 | 0.645 | 0.500 | 0.527 | 0.800 |
| exp\|M2 | 0.874 | 0.801 | 0.800 | 0.648 | 0.444 | 0.613 | 0.805 |
| imp\|M1 | 0.859 | 0.790 | 0.773 | 0.626 | 0.266 | 0.416 | 0.778 |
| imp\|M2 | 0.881 | 0.818 | 0.800 | 0.648 | 0.525 | 0.596 | 0.813 |
| char\|M1 | 0.866 | 0.794 | 0.791 | 0.642 | 0.441 | 0.544 | 0.796 |
| char\|M2 | 0.884 | 0.815 | 0.805 | 0.722 | 0.456 | 0.625 | **0.820** |

All the three kinds of features perform well on Data #3. Character features perform better than explicit expression and implicit expression. It can be concluded that character features are more suitable for short texts in our models, compared with the results on Data #1 and Data #2.

The three datasets have their own distinguishable characteristics. Among them, news titles are the shortest and the most coherent. The length, content and style of news comments are free and abundant. News bodies are the longest, with fixed format and rich content. Explicit, implicit and character features on the three datasets in this paper show different capabilities on calculating the sentiment orientation in different datasets. The detailed results are shown in Table 7. From the table, character features perform competitively on news titles and news comments with that on news bodies. It means character features perform well on short texts, but bad on long texts in our models. The results of explicit expression and implicit expression are relatively stable.

## 4.4 Analysis

This section focuses on the analysis of inter-sentiment correlation. The confusion matrix is employed to represent $C(e_a|e_b, D_i, F_j, M_k)$ in eq. 13. Instead of setting the fixed value of the threshold $\theta$, a dynamic $\theta$ is set in the analysis as selecting the top confusing sentiment pairs.

The confusion matrix of Data #1 is illustrated in Fig. 12. The models are likely to be confused between the sentiments *sadness* (ng), *love* (gd), and *anger* (fn). Specifically, *sadness* (ng) is likely to be recognized as *anger* (fn) or *love* (gd). *Love* (gd) is likely to be judged as *anger* (fn). By contrast, *anger* (fn) is unlikely to be judged as *fear* (zj) or *surprise* (xq). In short, comments can be easily mistaken as *anger*, even though they are not.

The confusion matrix of Data #2 and Data #3 are shown in Fig. 13 and Fig. 14. The two datasets are both objective data, which are used to observe the inter-sentiment correlation. It can be seen from the figures that the confusion degree of news bodies is higher than that of news titles, which means that it is harder to recognize the sentiments aroused by news bodies than news titles. Based on the voting result, the objective contents that cause *sadness* (ng) and *anger* (fn) are easy to be misjudged as *love* (gd) and *joy* (gx).

If our models are a person who has sentiment cognitive bias, then the above results can explain and estimate how people misunderstand other's words. In subjective texts, it is likely to misinterpret the sentiment in one's words as *anger* (fn), rather than *fear* (zj) or *surprise* (xq). If the sentiment in one's words is actually *sadness* (ng), it is more likely to be judged as other sentiments incorrectly. In objective texts, if a sample text causes *anger* (fn), the models are likely to predict that it will cause *love* (gd), but not likely to cause *surprise* (xq).

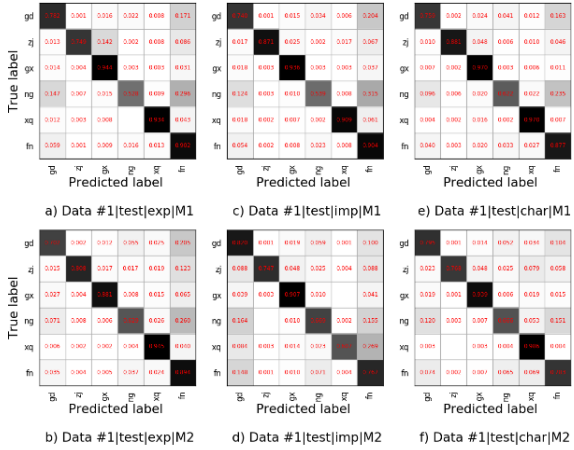

**Figure 12: Confusion matrix on test set of Data #1**

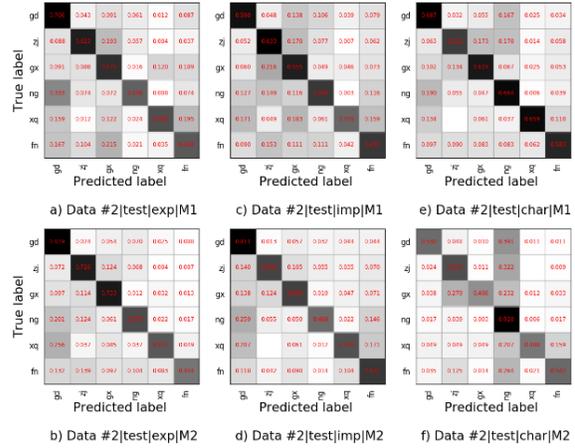

**Figure 13: Confusion matrix on test set of Data #2**

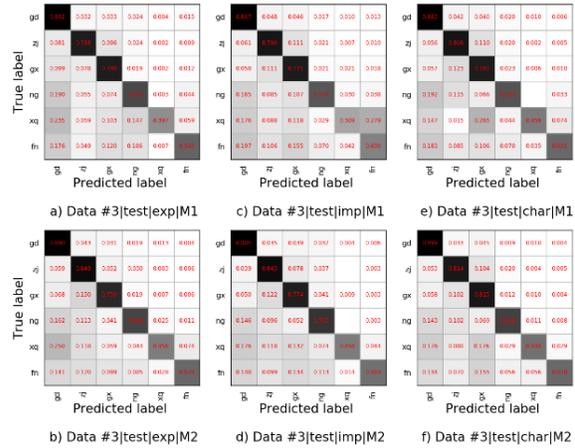

**Figure 14: Confusion matrix on test set of Data #3**

There is a controversial phenomenon on the interpretation of *anger* (fn) and *love* (gd). In subjective news comments, a text is likely to be considered as *anger*. By contrast, in objective news bodies and titles, it is easy to predict that a text may cause *love* (gd). One of the reasons for this phenomenon may be the fact that the news articles must be written without prejudice. Therefore, it is not easy to predict *anger* from the news articles through its text features. However, the readers are free to express their feelings. The netizen's sentiments are easy to be transferred to *anger* (fn), no matter what sentiment they want to express. Thus, as a comment conversation keeps going on, it is likely to end up with *anger*.

## 5. CONCLUSIONS

Mining the inter-sentiment correlations in web events is significant to track its development. The sentiments, though caused by objective information, are subjective, compound, and diverse, which makes the sentiment recognition hard. This paper employs six basic kinds of emotions – *love, joy, anger, sadness, fear, and surprise* – to analyse the sentiment in texts. Three kinds of features and two deep neural network models are proposed and applied to three datasets.

The two deep neural network models are presented for sentiment calculation. Three datasets are collected, covering both objective and subjective texts in varying lengths (long and short). In terms of the subjective comment data, the sentiment *sadness* is hard to be recognized, which is likely to be misinterpreted to other sentiments. The reason might be the overlap of the words or characters between the sentiment *sadness* and the others. All the sentiments are likely to be recognized as *anger*. In terms of the objective information, the text which arouses *anger* is relatively unpredictable and likely to be classified to *love*.

The controversial phenomenon on the interpretation of *anger* (fn) and *love* (gd) also draws our attention. In subjective text, other emotions can easily be considered as *anger*. By contrast, in objective news bodies and titles, it is easy to regard text as caused *love* (gd). It means, journalist may want to arouse emotion *love* by writing news, but cause *anger* after the news has been published. This result reflects the sentiment complexity and unpredictability.

## Acknowledgement

This work is supported by National Key R&D Program of China (No. 2017YFC0803300), National Science Foundation of China (Grant Nos. 91646201, U1633203), and CSC(China Scholarship Council).

## 6. REFERENCES

[1] Mishra, A., Dey, K., & Bhattacharyya, P. (2017). Learning cognitive features from gaze data for sentiment and sarcasm classification using convolutional neural network. In Proceedings of the 55th Annual Meeting of the Association for Computational Linguistics (Volume 1: Long Papers) (Vol. 1, pp. 377-387).
[2] Peled, L., & Reichart, R. (2017). Sarcasm SIGN: Interpreting Sarcasm with Sentiment Based Monolingual Machine Translation. arXiv preprint arXiv:1704.06836.
[3] Gui, X., Wang, Y., Kou, Y., Reynolds, T. L., Chen, Y., & Mei, Q., et al. (2017). Understanding the Patterns of Health Information Dissemination on Social Media during the Zika Outbreak. AMIA 2017 Symposium.
[4] Liang, S., Yilmaz, E., & Kanoulas, E. (2016, August). Dynamic clustering of streaming short documents. In Proceedings of the 22nd ACM SIGKDD International Conference on Knowledge Discovery and Data Mining (pp. 995-1004). ACM.
[5] Liang, S., Ren, Z., Yilmaz, E., & Kanoulas, E. (2017). Collaborative User Clustering for Short Text Streams. In AAAI(pp. 3504-3510).
[6] Zhao, Y., Liang, S., Ren, Z., Ma, J., Yilmaz, E., & de Rijke, M. (2016, July). Explainable user clustering in short text streams. In Proceedings of the 39th International ACM SIGIR conference on Research and Development in Information Retrieval (pp. 155-164). ACM.
[7] Xiao, L., Min, Z., Yiqun, L., & Shaoping, M. (2017, November). A Neural Network Model for Social-Aware Recommendation. In Asia Information Retrieval Symposium (pp. 125-137). Springer, Cham.
[8] Mao, J., Liu, Y., Luan, H., Zhang, M., Ma, S., Luo, H., & Zhang, Y. (2017, August). Understanding and Predicting Usefulness Judgment in Web Search. In Proceedings of the 40th International ACM SIGIR Conference on Research and Development in Information Retrieval (pp. 1169-1172). ACM.
[9] Mao, J., Liu, Y., Zhou, K., Nie, J. Y., Song, J., Zhang, M., ... & Luo, H. (2016, July). When does Relevance Mean Usefulness and User Satisfaction in Web Search?. In Proceedings of the 39th International ACM SIGIR conference on Research and Development in Information Retrieval (pp. 463-472). ACM.
[10] Mikolov, T., Sutskever, I., Chen, K., Corrado, G. S., & Dean, J. (2013). Distributed representations of words and phrases and their compositionality. In Advances in neural information processing systems (pp. 3111-3119).
[11] Pennington, J., Socher, R., & Manning, C. (2014). Glove: Global vectors for word representation. In Proceedings of the 2014 conference on empirical methods in natural language processing (EMNLP) (pp. 1532-1543).
[12] Kuang, S., & Davison, B. D. (2017). Learning Word Embeddings with Chi-Square Weights for Healthcare Tweet Classification. Applied Sciences, 7(8), 846.
[13] Das, S. R., & Chen, M. Y. (2007). Yahoo! for Amazon: Sentiment extraction from small talk on the web. Management science, 53(9), 1375-1388.
[14] Rao, Y., Lei, J., Wenyin, L., Li, Q., & Chen, M. (2014). Building emotional dictionary for sentiment analysis of online news. World Wide Web, 17(4), 723-742.
[15] Wang, L., & Xia, R. (2017). Sentiment Lexicon Construction with Representation Learning Based on Hierarchical Sentiment Supervision. In Proceedings of the 2017 Conference on Empirical Methods in Natural Language Processing (pp. 502-510).
[16] Luo, X., Xu, Z., Yu, J., & Chen, X. (2011). Building association link network for semantic link on web resources. IEEE transactions on automation science and engineering, 8(3), 482-494.
[17] Huang, S., Niu, Z., & Shi, C. (2014). Automatic construction of domain-specific sentiment lexicon based on constrained label propagation. Knowledge-Based Systems, 56, 191-200.
[18] Qiu, G., Liu, B., Bu, J., & Chen, C. (2009, July). Expanding domain sentiment lexicon through double propagation. In IJCAI (Vol. 9, pp. 1199-1204).
[19] Ganapathibhotla, G., & Liu, B. (2008). Identifying preferred entities in comparative sentences. In Proceedings of the International Conference on Computational Linguistics, COLING.
[20] Zheng, M., Lei, Z., Liao, X., & CHEN, G. (2013). Identify sentiment-objects from Chinese sentences based on cascaded conditional random fields. J. Chin. Inf. Process, 27(3), 69-76.
[21] Zhu, Y., Tian, H., Ma, J., Liu, J., & Liang, T. (2014, October). An integrated method for micro-blog subjective sentence identification based on three-way decisions and naive bayes. In International Conference on Rough Sets and Knowledge Technology (pp. 844-855). Springer, Cham.
[22] Pang, B., & Lee, L. (2008). Opinion mining and sentiment analysis. Foundations and Trends® in Information Retrieval, 2(1–2), 1-135.
[23] Karamibekr, M., & Ghorbani, A. A. (2013, October). Lexical-syntactical patterns for subjectivity analysis of social issues. In International Conference on Active Media Technology (pp. 241-250). Springer, Cham.
[24]. Chung, J., Gulcehre, C., Cho, K., & Bengio, Y. (2014). Empirical evaluation of gated recurrent neural networks on sequence modeling. arXiv preprint arXiv:1412.3555.
[25] Qian, Q., Huang, M., Lei, J., & Zhu, X. (2016). Linguistically regularized lstms for sentiment classification. arXiv preprint arXiv:1611.03949.
[26]. Yoon Kim. 2014. Convolutional neural networks for sentence classification. In EMNLP. pages 1746– 1751
[27]. Blunsom, P., Grefenstette, E., & Kalchbrenner, N. (2014). A convolutional neural network for modelling sentences. Proceedings of the 52nd Annual Meeting of the Association for Computational Linguistics, ACL. pages 655–665
[28]. Richard Socher, Jeffrey Pennington, Eric H Huang, Andrew Y Ng, and Christopher D Manning. 2011. Semi-supervised recursive autoencoders for predicting sentiment distributions. In EMNLP. pages 151– 161.
[29]. Richard Socher, Alex Perelygin, Jean Y Wu, Jason Chuang, Christopher D Manning, Andrew Y Ng, and Christopher Potts. 2013. Recursive deep models for semantic compositionality over a sentiment treebank. In EMNLP. pages 1631–1642.
[30] Liu, P., Qiu, X., & Huang, X. (2017). Adversarial Multi-task Learning for Text Classification. arXiv preprint arXiv:1704.05742.
[31] Chen, P., Sun, Z., Bing, L., & Yang, W. (2017). Recurrent Attention Network on Memory for Aspect Sentiment Analysis. In Proceedings of the 2017 Conference on Empirical Methods in Natural Language Processing (pp. 463-472).
[32] Long, Y., Lu, Q., Xiang, R., Li, M., & Huang, C. R. (2017). A Cognition Based Attention Model for Sentiment Analysis. Conference on Empirical Methods in Natural Language Processing.
[33] Wilson, T., Wiebe, J., & Hwa, R. (2004, July). Just how mad are you? Finding strong and weak opinion clauses. In aaai (Vol. 4, pp. 761-769).
[34] Parrott, W. G. (Ed.). (2001). Emotions in social psychology: Essential readings. Psychology Press.